\pdfoutput=1

\documentclass[11pt]{article}

\usepackage{acl}

\usepackage{times}
\usepackage{latexsym}

\usepackage{listings}
\usepackage{amssymb}
\usepackage{amsmath}

\DeclareMathOperator*{\argmax}{arg\,max}

\usepackage[T1]{fontenc}

\usepackage[utf8]{inputenc}

\usepackage{microtype}

\usepackage[pdftex]{graphicx}

\title{Jam or Cream First?$^1$ Modeling Ambiguity in Neural Machine Translation with SCONES}

\author{Felix Stahlberg \and Shankar Kumar \\
        Google Research \\ \texttt{\{fstahlberg,shankarkumar\}@google.com}}

\begin{document}
\maketitle
\begin{abstract}
The softmax layer in neural machine translation is designed to model the distribution over mutually exclusive tokens. Machine translation, however, is intrinsically uncertain: the same source sentence can have multiple semantically equivalent translations. Therefore, we propose to replace the softmax activation with a multi-label classification layer that can model ambiguity more effectively. We call our loss function Single-label Contrastive Objective for Non-Exclusive Sequences (SCONES). We show that the multi-label output layer can still be trained on single reference training data using the SCONES loss function. SCONES yields consistent BLEU score gains across six translation directions, particularly for medium-resource language pairs and small beam sizes. By using smaller beam sizes we can speed up inference by a factor of 3.9x and still match or improve the BLEU score obtained using softmax. Furthermore, we demonstrate that SCONES can be used to train NMT models that assign the highest probability to adequate translations, thus mitigating the ``beam search curse''. Additional experiments on synthetic language pairs with varying levels of uncertainty suggest that the improvements from SCONES can be attributed to better handling of ambiguity.
\end{abstract}

\section{Introduction}

\setcounter{footnote}{1}
\footnotetext{\url{https://en.wikipedia.org/wiki/Cream_tea\#Variations}}

Conventional neural machine translation (NMT) models learn the probability $P(\mathbf{y}|\mathbf{x})$ of the target sentence $\mathbf{y}$ given the source sentence $\mathbf{x}$ \citep{kalchbrenner-blunsom-2013-recurrent,sutskever-seq2seq}. This framework implies that there is a single best translation for a given source sentence: if there were multiple valid translations $\mathbf{y}_1$ and $\mathbf{y}_2$ they would need to share probability mass (e.g. $P(\mathbf{y}_1|\mathbf{x})=0.5$ and $P(\mathbf{y}_2|\mathbf{x})=0.5$), but such a distribution could also represent {\em model} uncertainty, i.e.\ the case when {\em either} $\mathbf{y}_1$ {\em or} $\mathbf{y}_2$ are correct translations. Therefore, learning a single distribution over all target language sentences does not allow the model to naturally express {\em intrinsic uncertainty}\footnote{This is sometimes referred to as {\em aleatoric} uncertainty in the literature \citep{ml-uncertainty}.} \citep{mt-uncertainty,dreyer-marcu-2012-hyter,nmt-uncertainty,seq2seq-tractability}, the nature of the translation task to allow multiple semantically equivalent translations for a given source sentence. A single distribution over all sequences represents uncertainty by assigning probabilities, but it cannot distinguish between different kinds of uncertainty (e.g.\ model uncertainty versus intrinsic uncertainty).

Therefore, in this work we frame machine translation as a multi-label classification task~\cite{tsoumakas07multilabelclassification,zhang14multilabel}. Rather than learning a single distribution $P(\mathbf{y}|\mathbf{x})$ over all target sentences $\mathbf{y}$ for a source sentence $\mathbf{x}$, we learn binary classifiers for each sentence pair $(\mathbf{x},\mathbf{y})$ that indicate whether or not $\mathbf{y}$ is a valid translation of $\mathbf{x}$. In this framework, intrinsic uncertainty can be represented by setting the probabilities of two (or more) correct translations $\mathbf{y}_1$ and $\mathbf{y}_2$ to 1 simultaneously. The probabilities for each translation are computed using separate binary classifiers, and thus there is no requirement that the probabilities sum to one over all translations. 
In practice, the probability of a complete translation is decomposed into a product of the token-level probabilities. Thus we replace the softmax output layer in Transformer  models \citep{transformer} with sigmoid activations that assign a probability between 0 and 1 to each token in the vocabulary at each time step. We propose a loss function, \emph{\textbf{S}ingle-label \textbf{C}ontrastive \textbf{O}bjective for \textbf{N}on-\textbf{E}xclusive \textbf{S}equences} (\textbf{SCONES}) that allows us to train our models on single reference training data. Our work is inspired by noise-contrastive estimation (NCE) \citep{nce, nce-lm}. Unlike NCE, whose primary goal was to efficiently train models over large vocabularies, our motivation for SCONES is to model non-exclusive outputs.

\begin{figure*}[t!]
\centering
\small
\includegraphics[width=0.75\textwidth]{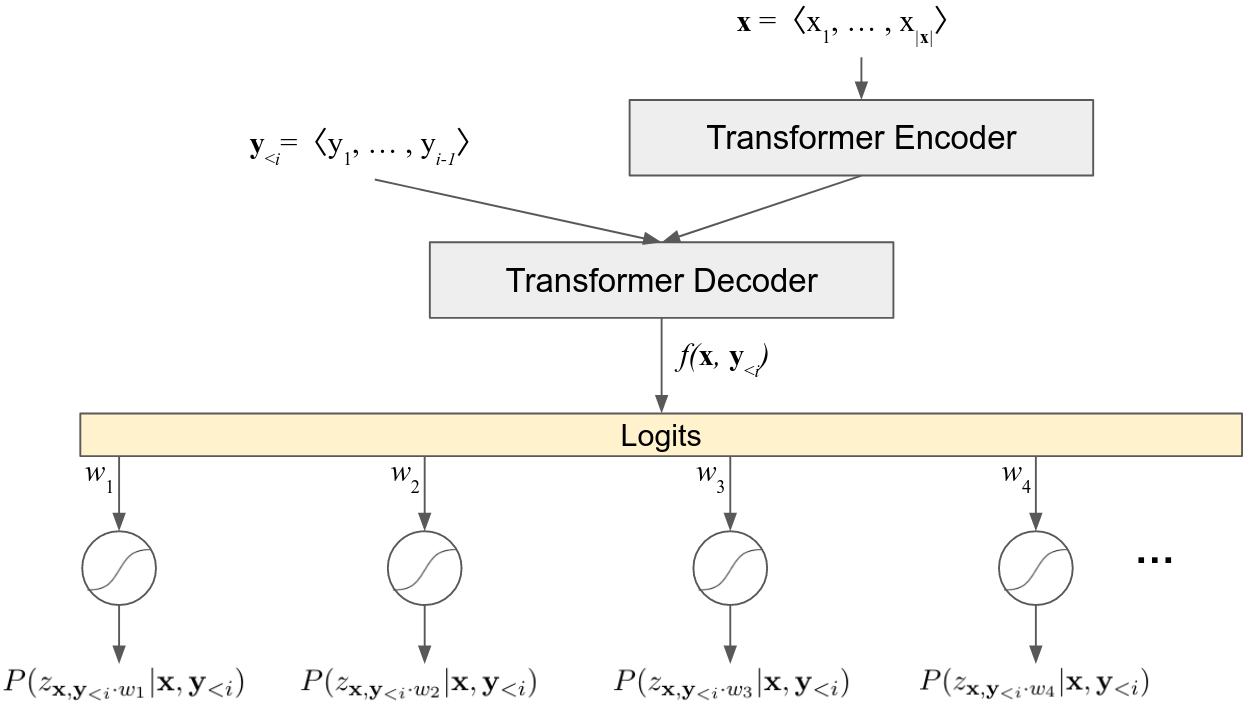}
\caption{Multi-way NMT Transformer architecture for non-exclusive target sequences.}
\label{fig:architecture}
\end{figure*}

We demonstrate multiple benefits of training NMT models using SCONES when compared to standard cross-entropy with regular softmax. We report consistent BLEU score gains between 1\%-9\% across six different translation directions. SCONES with greedy search typically outperforms softmax with beam search, resulting in inference speed-ups of up to 3.9x compared to softmax without any degradation in BLEU score.

SCONES can be tuned to mitigate some of the pathologies of traditional NMT models. Softmax-based models have been shown to assign the highest probability to either empty or inadequate translations (modes) \citep{stahlberg-byrne-2019-nmt,eikema-aziz-2020-map}. This behavior manifests itself as the ``beam search curse'' \citep{koehn-knowles-2017-six}: increasing the beam size may lead to worse translation quality. We show that SCONES can be used to train models that a) assign the highest probability to adequate translations and b) do not suffer from the beam search curse. 

Finally, we use SCONES to train models on synthetic translation pairs that we generate by sampling from the IBM Model 3 \citep{brown-etal-1993-mathematics}. By varying the sampling temperature, we control the level of ambiguity in the language pair. We show that SCONES is effective in improving the adequacy of the highest probability translation for highly ambiguous translation pairs, confirming our intuition that SCONES can handle intrinsic uncertainty well.

\section{Training NMT models with SCONES}
\label{sec:scones}

We denote the (subword) vocabulary as $\mathcal{V}=\{w_1,\dots,w_{|\mathcal{V}|}\}$, the special end-of-sentence symbol as $w_1=\text{</s>}$, the source sentence as $\mathbf{x}=\langle x_1, \dots, x_{|\mathbf{x}|} \rangle\in\mathcal{V}^*$, a translation as $\mathbf{y}=\langle y_1, \dots, y_{|\mathbf{y}|} \rangle\in\mathcal{V}^*$, and a translation prefix as $\mathbf{y}_{\leq i}=\langle y_1, \dots, y_i\rangle$. We use a center dot ``$\cdot$'' for string concatenations. Unlike conventional NMT that models a single distribution $P(\mathbf{y}|\mathbf{x})$ over all target language sentences, SCONES learns a separate binary classifier for each sentence pair $(\mathbf{x},\mathbf{y})$.  We define a Boolean function $t(\cdot, \cdot)$ that indicates whether $\mathbf{y}$ is a valid translation of $\mathbf{x}$:
\begin{equation}
  t(\mathbf{x},\mathbf{y}) :=
    \begin{cases}
      \mathtt{true} & \text{if $\mathbf{y}$ is a translation of $\mathbf{x}$}\\
      \mathtt{false} & \text{otherwise}
    \end{cases}.
\end{equation}
We do not model $t(\cdot,\cdot)$ directly. To guide decoding, we learn variables $z_{\mathbf{x},\mathbf{y}}$ which generalize $t(\cdot,\cdot)$ to translation {\em prefixes}:
\begin{equation}
  z_{\mathbf{x},\mathbf{y}} :=
    \begin{cases}
      1 & \exists \mathbf{y}'\in \mathcal{V}^*:  t(\mathbf{x}, \mathbf{y} \cdot \mathbf{y}') = \mathtt{true} \\
      0 & \text{otherwise}
    \end{cases},
\end{equation}
i.e. $z_{\mathbf{x},\mathbf{y}}$ is a binary label for the pair $(\mathbf{x}, \mathbf{y})$ consisting of source sentence $\mathbf{x}$ and the translation prefix $\mathbf{y}$: $z_{\mathbf{x},\mathbf{y}}=1$ iff.\ $\mathbf{y}$ is a prefix of a valid translation of $\mathbf{x}$. We decompose its probability as a product of conditionals to facilitate left-to-right beam decoding:\footnote{As a base case we define $P(z_{\mathbf{x},\epsilon}=1|\mathbf{x}) = 1$ for the empty translation prefix.}
\begin{equation}
\begin{aligned}
P(z_{\mathbf{x},\mathbf{y}}=1|\mathbf{x}) &:=& \mkern-12mu\prod_{i=1}^{|\mathbf{y}|} P(z_{\mathbf{x},\mathbf{y}_{\leq i}}\mkern-2mu =1|z_{\mathbf{x},\mathbf{y}_{< i}}\mkern-2mu =1, \mathbf{x}) \\
&=& \prod_{i=1}^{|\mathbf{y}|} P(z_{\mathbf{x},\mathbf{y}_{\leq i}}\mkern-2mu =1|\mathbf{x}, \mathbf{y}_{< i}).
\end{aligned}
\label{eq:fact}
\end{equation}
We assign the conditional probabilities by applying the sigmoid activation function $\sigma(\cdot)$ to the logits:
\begin{equation}
P(z_{\mathbf{x},\mathbf{y}_{< i}\cdot w}=1|\mathbf{x}, \mathbf{y}_{< i}) = \sigma(f(\mathbf{x}, \mathbf{y}_{< i})_w),
\label{eq:logits}
\end{equation}
where $w\in\mathcal{V}$ is a single token, $f(\mathbf{x}, \mathbf{y}_{< i})\in \mathbb{R}^{|\mathcal{V}|}$ are the logits at time step $i$, and $f(\mathbf{x}, \mathbf{y}_{< i})_{w}$ is the logit corresponding to token $w$. The only architectural difference to a standard NMT model is the output activation: instead of the softmax function that yields a single distribution over the full vocabulary, we use multiple sigmoid activations in each logit component to define separate Bernoulli distributions for each item in the vocabulary (Fig.\ \ref{fig:architecture}). However, using  such a multi-label classification view requires a different training loss function because, unlike the probabilities from a softmax, the probabilities in Eq.~\ref{eq:logits} do not provide a normalized distribution over the vocabulary. An additional challenge is that existing MT training datasets typically do not provide more than one reference translation. Our SCONES loss function aims to balance two token-level objectives using a scaling factor $\alpha\in\mathbb{R}^+$:
\begin{equation}
\mathcal{L}(\mathbf{x},\mathbf{y})=\frac{1}{|\mathbf{y}|}\sum_{i=1}^{|\mathbf{y}|} \mathcal{L}_\text{SCONES}(\mathbf{x},\mathbf{y}, i),
\end{equation}
where
\begin{equation}
\mathcal{L}_\text{SCONES}(\mathbf{x},\mathbf{y}, i) = \mathcal{L}_+(\mathbf{x},\mathbf{y}, i) + \alpha\mathcal{L}_-(\mathbf{x},\mathbf{y}, i).
\label{eq:scones}
\end{equation}
$\mathcal{L}_+(\cdot)$ aims to increase the probability $P(z_{\mathbf{x},\mathbf{y}_{\leq i}}=1|\mathbf{x}, \mathbf{y}_{< i})$ of the gold label $y_i$ since it is a valid extension of the translation prefix $\mathbf{y}_{<i}$:
\begin{equation}
\begin{aligned}
\mathcal{L}_+(\mathbf{x},\mathbf{y}, i) &=& -\log P(z_{\mathbf{x},\mathbf{y}_{\leq i}}=1|\mathbf{x}, \mathbf{y}_{< i}) \\
  &=& -\log \sigma(f(\mathbf{x}, \mathbf{y}_{< i})_{y_i}).
\end{aligned}
\end{equation}
$\mathcal{L}_-(\cdot)$ is designed to reduce the probability $P(z_{\mathbf{x},\mathbf{y}_{< i}\cdot w}=1|\mathbf{x}, \mathbf{y}_{< i})$ for all labels $w$ except for the gold label $y_i$:
\begin{equation}
\begin{aligned}
\mathcal{L}_-(\mathbf{x},\mathbf{y}, i) &=& \mkern-12mu - \mkern-14mu \sum_{w\in \mathcal{V}\setminus\{y_i\}} \mkern-14mu &\log P(z_{\mathbf{x},\mathbf{y}_{< i}\cdot w}=0|\mathbf{x}, \mathbf{y}_{< i}) \\
&=& \mkern-12mu - \mkern-14mu \sum_{w\in \mathcal{V}\setminus\{y_i\}} \mkern-14mu &\log (1 - \sigma(f(\mathbf{x}, \mathbf{y}_{< i})_w)).
\end{aligned}
\end{equation}
Appendix \ref{sec:jax} provides an implementation of SCONES in JAX \citep{jax}.

During inference we search for the translation $\mathbf{y}^*$ that ends with $\text{</s>}$ and has the highest probability of being a translation of $\mathbf{x}$:
\begin{equation}
\begin{aligned}
\mathbf{y}^* &=& \mkern-14mu \argmax_{\mathbf{y}\in\{\mathbf{w}\cdot\text{</s>}|\mathbf{w}\in\mathcal{V}^*\}} & P(z_{\mathbf{x},\mathbf{y}}=1|\mathbf{x}) \\
&\overset{\text{Eqs.\ \ref{eq:fact}, \ref{eq:logits}}}{=}& \mkern-14mu \argmax_{\mathbf{y}\in\{\mathbf{w}\cdot\text{</s>}|\mathbf{w}\in\mathcal{V}^*\}} & \sum_{i=1}^{|\mathbf{y}|} \log\sigma(f(\mathbf{x}, \mathbf{y}_{< i})_{y_i}).
\end{aligned}
\end{equation}
We approximate this decision rule with vanilla beam search. The same inference code is used for both our softmax baselines and the SCONES-trained models. The only difference is that the logits from SCONES models are transformed by a sigmoid instead of a softmax activation, i.e.\ no summation over the full vocabulary is necessary.

\begin{table}[t!]
\centering
\small
\begin{tabular}{ll}
\hline \textbf{Parameter} & \textbf{Value} \\ \hline
Attention dropout rate & 0.1 \\
Attention layer size & 512 \\
Dropout rate & 0.1 \\
Embedding size & 512 \\	
MLP dimension & 2,048 \\	
Number of attention heads & 8 \\
Number of layers & 6 \\
Training batch size & 256 \\
Total number of parameters & 121M \\
\hline
\end{tabular}
\caption{\label{tab:trans-hyper} Transformer hyper-parameters.}
\end{table}

\begin{table}[t!]
\centering
\small
\begin{tabular}{lll}
\hline \textbf{Language pair} & \multicolumn{2}{c}{\textbf{\#Training sentence pairs}} \\
 & \textbf{Unfiltered} & \textbf{Filtered} \\ \hline
German-English & 39M & 33M \\
Finnish-English & 6.6M & 5.5M \\
Lithuanian-English & 2.3M & 2.0M \\
\hline
\end{tabular}
\caption{\label{tab:mt-train-set} MT training set sizes.}
\end{table}


\begin{table*}[t!]
\centering
\small
\begin{tabular}{l|c@{\hspace{1.1em}}c@{\hspace{1.1em}}c@{\hspace{1.1em}}c@{\hspace{1.1em}}c@{\hspace{1.1em}}c|c@{\hspace{1.1em}}c@{\hspace{1.1em}}c@{\hspace{1.1em}}c@{\hspace{1.1em}}c@{\hspace{1.1em}}c}
\hline  
& \multicolumn{6}{c|}{\textbf{Greedy search}} & \multicolumn{6}{c}{\textbf{Beam search (beam size = 4)}} \\
&  \textbf{de-en} & \textbf{en-de} & \textbf{fi-en} & \textbf{en-fi} & \textbf{lt-en} &\textbf{en-lt} &  \textbf{de-en} & \textbf{en-de} & \textbf{fi-en} & \textbf{en-fi} & \textbf{lt-en} &\textbf{en-lt} \\
\hline
Softmax & 38.8 & 38.7 & 26.9 & 18.5 & 26.3 & 11.5 & 39.6 & 39.4 & 27.7 & 19.0 & 26.9 & 12.0 \\
SCONES & 39.9 & 39.1 & 27.6 & 19.5 & 27.7 & 12.5 & 40.3 & 39.8 & 28.4 & 20.0 & 28.9 & 12.6\\
\hline
\textbf{Rel.\ improvement} & \textbf{+2.7}$^\ddagger$ & \textbf{+1.2} & \textbf{+2.8}$^\dagger$ & \textbf{+5.4}$^\ddagger$ & \textbf{+5.3}$^\ddagger$ & \textbf{+8.5}$^\ddagger$ & \textbf{+1.7}$^\dagger$ & \textbf{+0.9} & \textbf{+2.7}$^\dagger$ & \textbf{+5.5}$^\ddagger$ & \textbf{+7.4}$^\ddagger$ & \textbf{+5.7} \\
\end{tabular}
\caption{\label{tab:comparison-softmax-scones} BLEU score gains from SCONES over our NMT softmax baselines with tuned $\alpha$-values (Table \ref{tab:alpha}). Using a paired bootstrap method \citep{koehn-2004-statistical}, we highlight improvements that are statistically significant either at a .05 level ($\dagger$) or a .01 level ($\ddagger$).}
\end{table*}

\begin{table*}[t!]
\centering
\small
\begin{tabular}{l|c@{\hspace{1.1em}}c@{\hspace{1.1em}}c@{\hspace{1.1em}}c@{\hspace{1.1em}}c@{\hspace{1.1em}}c|c@{\hspace{1.1em}}c@{\hspace{1.1em}}c@{\hspace{1.1em}}c@{\hspace{1.1em}}c@{\hspace{1.1em}}c}
\hline  
& \multicolumn{6}{c|}{\textbf{Greedy search}} & \multicolumn{6}{c}{\textbf{Beam search (beam size = 4)}} \\
&  \textbf{de-en} & \textbf{en-de} & \textbf{fi-en} & \textbf{en-fi} & \textbf{lt-en} &\textbf{en-lt} &  \textbf{de-en} & \textbf{en-de} & \textbf{fi-en} & \textbf{en-fi} & \textbf{lt-en} &\textbf{en-lt} \\
\hline
Softmax &  70.44 & 68.08 & 68.93 & 66.16 & 68.52 & 56.68 & 70.78 & 68.48 & 69.56 & 66.44 & 69.20 & 57.61 \\
SCONES &  70.69 & 67.55 & 69.28 & 67.32 & 68.96 & 58.68 & 70.88 & 67.99 & 69.72 & 67.91 & 69.95 & 59.48 \\
\hline
\end{tabular}
\caption{\label{tab:comparison-softmax-scones-bleurt} BLEURT \citep{sellam-etal-2020-bleurt} scores (\texttt{BLEURT-20} checkpoint) for SCONES and our NMT softmax baselines with tuned $\alpha$-values (Table \ref{tab:alpha}).}
\end{table*}

\begin{table}[t!]
\centering
\small
\begin{tabular}{lc}
\hline \textbf{Language pair} & $\alpha$ \\
\hline
de-en & 0.5 \\
en-de & 0.5 \\
fi-en & 0.7 \\
en-fi & 1.0 \\
lt-en & 0.7 \\
en-lt & 0.9 \\
\hline
\end{tabular}
\caption{\label{tab:alpha} Values of $\alpha$ that yield the best greedy BLEU scores on the respective development sets.}
\end{table}

\paragraph{Relation to noise-contrastive estimation}

Our SCONES loss function is related to noise-contrastive estimation (NCE) \citep{nce,nce-lm} because both methods reformulate next word prediction as a multi-label classification problem, and both losses have a ``positive'' component for the gold label, and a ``negative'' component for other labels.\footnote{Technically, SCONES could be written as an instance of NCE with a scaling factor $\alpha$ and an exhaustive enumeration of negative NCE samples.} Unlike NCE, the negative loss component ($\mathcal{L}_-(\cdot)$) in SCONES does not require sampling from a noise distribution as it makes use of {\em all} tokens in the vocabulary besides the gold token. This is possible because we operate on a limited 32K subword vocabulary whereas NCE is typically used to efficiently train language models with much larger word-level vocabularies \citep{nce-lm}. NCE has a ``self-normalization'' property \citep{nce,nce-self-norm,nce-lm,goldberger-melamud-2018-self} which can reduce computation by avoiding the expensive partition function for distributions over the full vocabulary. To do so, NCE uses the multi-label classification task as a proxy problem. By contrast, in SCONES, the multi-label classification perspective is used to express the intrinsic uncertainty in MT and is not simply a proxy for the full softmax. Thus the primary motivation for SCONES is not self-normalization over the full vocabulary.

\section{Experimental setup}

In this work our focus is to compare NMT models trained with SCONES with well-trained standard softmax-based models. Thus we keep our setup simple, reproducible, and computationally economical. 
We trained Transformer models (Table~\ref{tab:trans-hyper}) in six translation directions -- German-English (de-en), Finnish-English (en-fi), Lithuanian-English (lt-en), and the reverse directions -- on the WMT19 \citep{barrault-etal-2019-findings} training sets as provided by TensorFlow Datasets.\footnote{\url{https://www.tensorflow.org/datasets/catalog/wmt19_translate}} We selected these language pairs to experiment with different training set sizes (Table \ref{tab:mt-train-set}). The training sets were filtered using language ID and simple length-based heuristics, and split into subwords using joint 32K SentencePiece \citep{kudo-richardson-2018-sentencepiece} models. 
All our models were trained until convergence on the development set (between 100K and 700K training steps) using the LAMB \citep{lamb} optimizer in JAX \citep{jax}. Our softmax baselines are trained by minimizing cross-entropy without label smoothing. Our multi-way NMT models are trained by minimizing the SCONES loss function from Sec.\ \ref{sec:scones}, also without label smoothing. We evaluate our models on the WMT19 test sets \citep{barrault-etal-2019-findings} with SacreBLEU \citep{post-2018-call},\footnote{Comparable to \url{http://wmt.ufal.cz/}.} using the WMT18 test sets as development sets to tune~$\alpha$.

\begin{figure*}[t!]
\centering
\small
\begin{tabular}{@{\hspace{0em}}c@{\hspace{0em}}c@{\hspace{0em}}}
\includegraphics[scale=1.0]{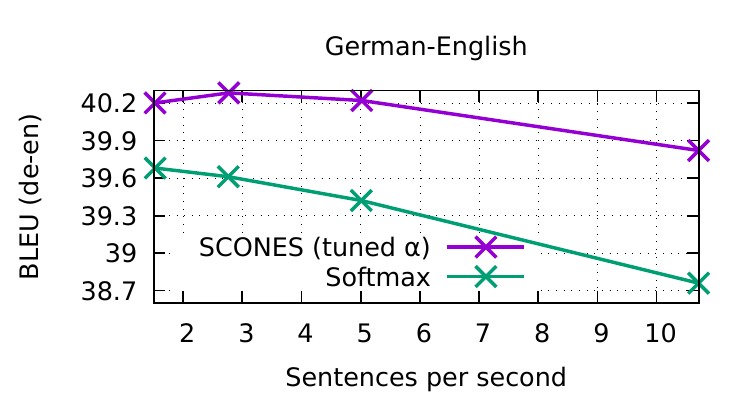} & \includegraphics[scale=1.0]{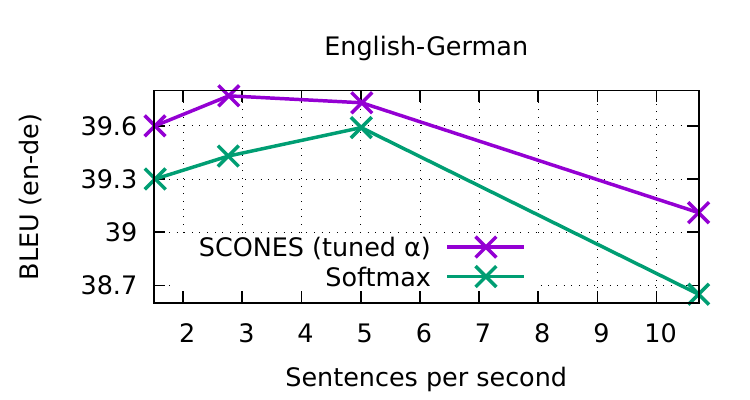} \\
\includegraphics[scale=1.0]{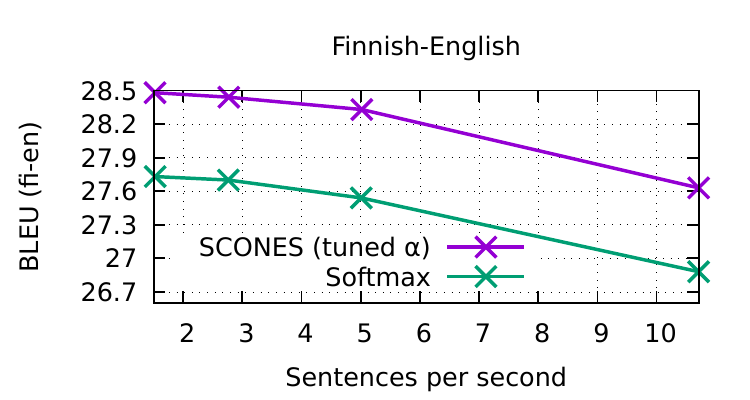} & \includegraphics[scale=1.0]{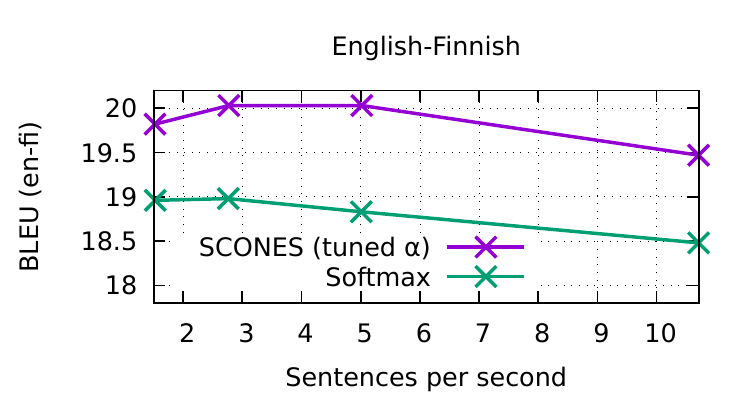} \\
\includegraphics[scale=1.0]{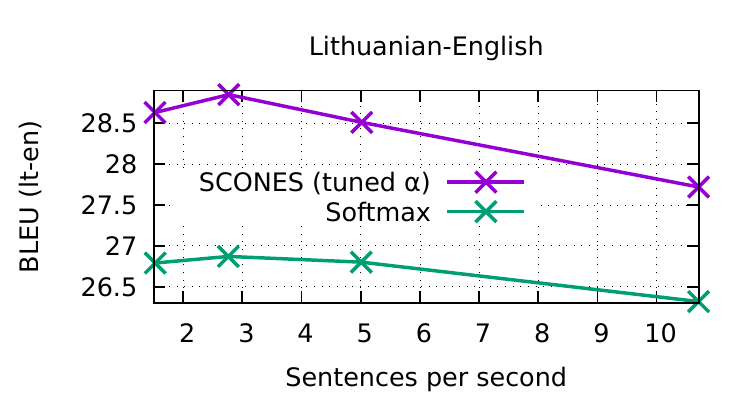} & \includegraphics[scale=1.0]{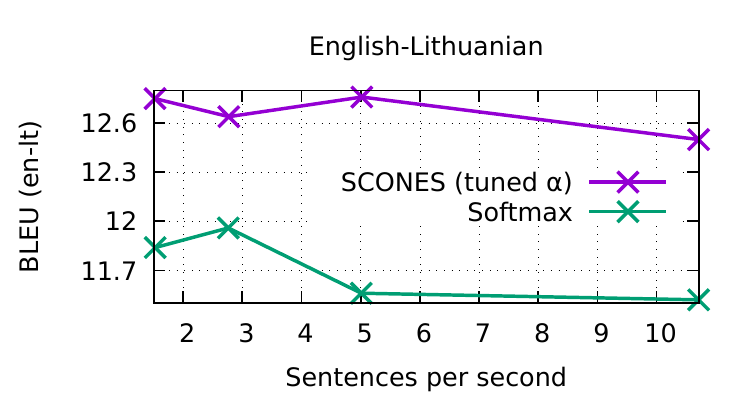} \\
\end{tabular}
\caption{BLEU scores as a function of GPU decoding speeds (median over five runs) for softmax and SCONES with greedy search and beam search with beam sizes 2, 4, and 8 (annotated with $\times$).}
\label{fig:runtime}
\end{figure*}

\section{Results}


\subsection{Translation quality}

Table \ref{tab:comparison-softmax-scones} compares our SCONES-based NMT systems with the softmax baselines when $\alpha$ is tuned based on the BLEU score on the development set (Table \ref{tab:alpha}). SCONES yields consistent improvements across the board. For four of six language pairs (all except en-de and fi-en), SCONES with greedy search is even able to outperform the softmax models with beam search. The language pairs with fewer resources (fi$\leftrightarrow$en, lt$\leftrightarrow$en) benefit from SCONES training much more than the high-resource language pairs (de$\leftrightarrow$en). SCONES still yields gains for all language directions except English-German when we use BLEURT \citep{sellam-etal-2020-bleurt} instead of BLEU as the evaluation measure (Table \ref{tab:comparison-softmax-scones-bleurt}).

\subsection{Decoding speed}

Our softmax-based models reach their (near) optimum BLEU score with a beam size of around 4. Most of our SCONES models can achieve similar or better BLEU scores with greedy search. Replacing beam-4 search with greedy search corresponds to a 3.9x speed-up (2.76 $\rightarrow$ 10.64 sentences per second) on an entry-level NVIDIA Quadro P1000 GPU with a batch size of 4.\footnote{As an additional optimization, our greedy search implementation operates directly on the logits without applying the output activations.} Fig.\ \ref{fig:runtime} shows the BLEU scores for all six translation directions as a function of decoding speed. Most of the speed-ups are due to choosing a smaller beam size and not due to SCONES avoiding the normalization over the full vocabulary. We expect further speed-ups when comparing models with larger vocabularies.

\subsection{Mitigating the beam search curse}

One of the most irksome pathologies of traditional softmax-based NMT models is the ``beam search curse'' \citep{koehn-knowles-2017-six}: larger beam sizes improve the log-probability of the translations, but the translation quality gets worse. This happens because with large beam sizes, the model prefers translations that are too short. This phenomenon has been linked to the local normalization in sequence models \citep{sountsov-sarawagi-2016-length,murray-chiang-2018-correcting} and poor model calibration \citep{nmt-calibration}. \citet{stahlberg-byrne-2019-nmt} showed that modes are often empty and suggested that the inherent bias of the model towards short translations is often obscured by beam search errors. \citet{seq2seq-tractability} provided strong evidence that this length deficiency is due to the intrinsic uncertainty of the MT task. Given that models trained with SCONES explicitly take into account inherent uncertainty, we ran an experiment to determine whether these models are more robust to the beam search curse compared to softmax trained models.

\begin{figure}[t!]
\centering
\small
\includegraphics[scale=1.0]{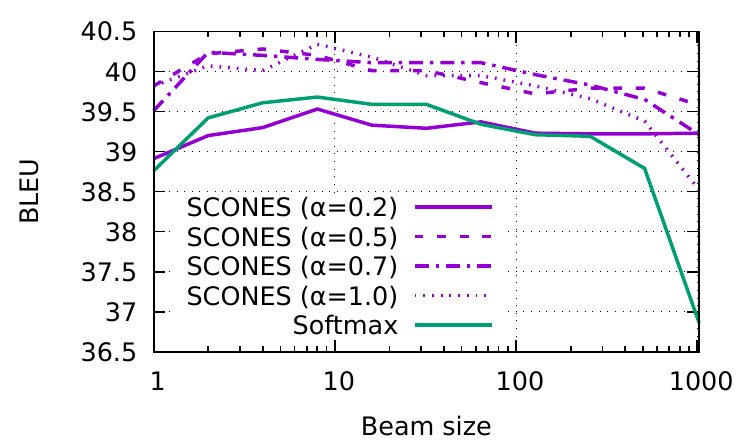}
\caption{German-English BLEU score as a function of beam size.}
\label{fig:deen_beam_size_bleu}
\end{figure}

\begin{figure}[t!]
\centering
\small
\includegraphics[scale=1.0]{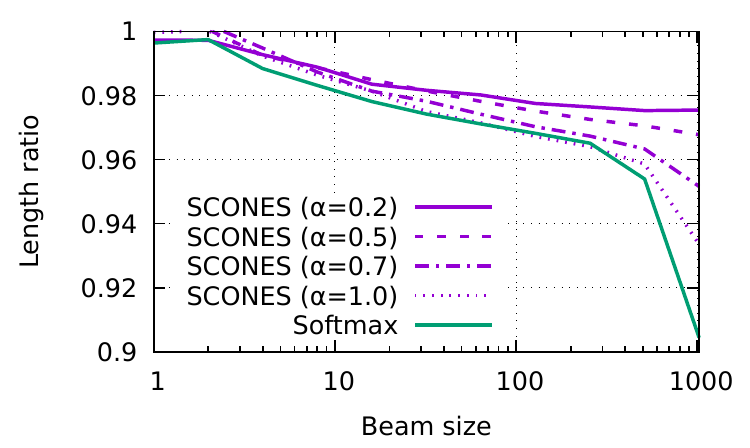}
\caption{German-English length ratio (hypothesis length / reference length) as a function of beam size.}
\label{fig:deen_beam_size_ratio}
\end{figure}

\begin{table*}
\centering
\small
\begin{tabular}{l|cccccc|cccccc}
\hline  
& \multicolumn{6}{c|}{\textbf{Beam search (beam size = 4)}} & \multicolumn{6}{c}{\textbf{Exact search}} \\
&  \textbf{de-en} & \textbf{en-de} & \textbf{fi-en} & \textbf{en-fi} & \textbf{lt-en} &\textbf{en-lt} &  \textbf{de-en} & \textbf{en-de} & \textbf{fi-en} & \textbf{en-fi} & \textbf{lt-en} &\textbf{en-lt} \\
\hline
Softmax & 39.6 & 39.4 & 27.7 & 19.0 & 26.9 & 12.0 & 23.7 & 15.6 & 16.7 & 10.1 & 14.2 & \ 7.1 \\
SCONES ($\alpha=0.2$) & 39.3 & 38.9 & 27.7 & 19.6 & 27.9 & 12.7 & 39.1& 37.2 & 26.7 & 18.7 & 25.6 & 12.1 \\
\hline
\end{tabular}
\caption{\label{tab:beam-vs-dfs} BLEU scores of beam search and exact search for all six translation directions.}
\end{table*}

\begin{figure}[t!]
\centering
\small
\includegraphics[scale=1.0]{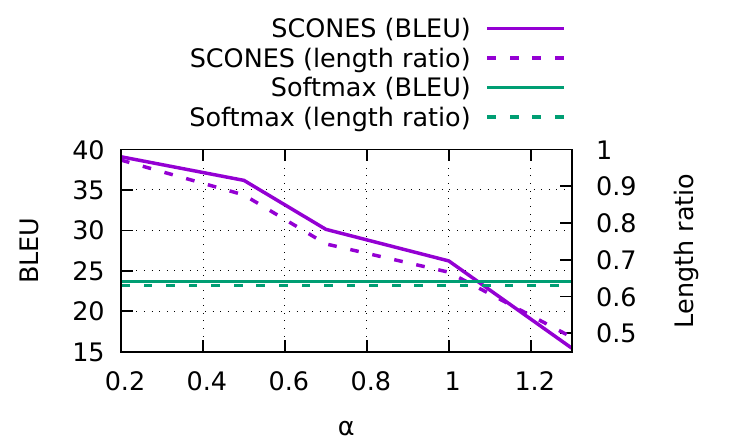}
\caption{German-English BLEU scores and length ratios (hypothesis length / reference length) for exact search.}
\label{fig:deen_exact_search}
\end{figure}

\begin{figure}[t!]
\centering
\small
\includegraphics[scale=1.0]{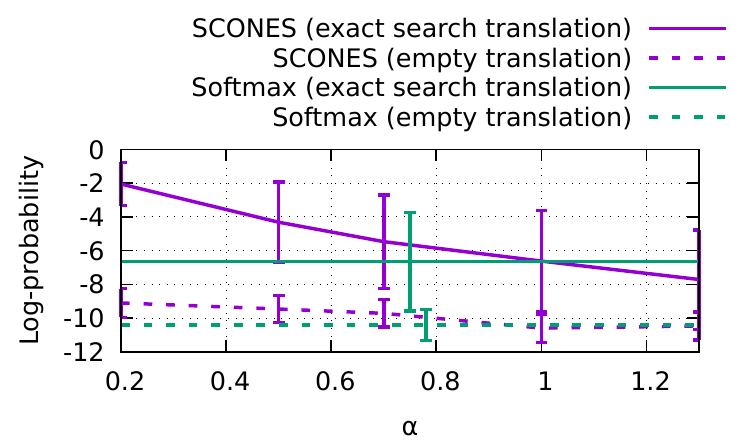}
\caption{Mean and standard deviation (error bars) of  log-probabilities of the global highest probability translations (found using exact search) and the empty translations for German-English.}
\label{fig:deen_logprob}
\end{figure}

Fig.\ \ref{fig:deen_beam_size_bleu} plots the BLEU score as a function of the beam size. The sharp decline of the green curve for large beam sizes reflects the \emph{beam search curse} for the softmax baseline. SCONES seems to be less affected at larger beam sizes, particularly for small $\alpha$-values: the BLEU score for SCONES with $\alpha=0.2$ (solid purple curve) is stable for beam sizes greater than 100. Fig.\ \ref{fig:deen_beam_size_ratio}, which displays the length ratio (the hypothesis length divided by the reference length) versus beam size, suggests that the differences in BLEU trajectories are partly due to translation lengths. Translations obtained using softmax become shorter at higher beam sizes whereas for SCONES with $\alpha=0.2$, there is no such steep decrease in length. To study the impact of $\alpha$ in the absence of beam search errors we ran the exact depth-first search algorithm of \citet{stahlberg-byrne-2019-nmt} to find the translation with global highest probability.\footnote{The maximum number of explored states per sentence was set to 1M. This threshold was reached for less than 1.45\% of the German-English sentences. See Appendix \ref{sec:dfs-maxedout} for other language directions.} The adequacy of the translations found by exact search depends heavily on $\alpha$ (Fig.\ \ref{fig:deen_exact_search}). With exact search, small $\alpha$-values yield adequate translations, but $\alpha\approx 1.0$ performs similar to the softmax baseline: the BLEU score drops because hypotheses are too short. Table \ref{tab:beam-vs-dfs} shows that SCONES with $\alpha = 0.2$ consistently outperforms the softmax baselines by a large margin with exact search. Fig.\ \ref{fig:deen_logprob} sheds some light on why SCONES with small $\alpha$ does not prefer empty translations. A small $\alpha$ leads to a larger gap between the log-probabilities of the exact search translation and the empty translation that arises from higher log-probabilities  for the exact-search translation along with smaller variances. Intuitively, a small $\alpha$ reduces the importance of the negative loss component $\mathcal{L}_-(\cdot)$ in Eq.\ \ref{eq:scones}, and thus biases each binary classifier towards predicting the \texttt{true} label.

\subsection{Reducing the number of beam search errors}

\begin{figure}[t!]
\centering
\small
\includegraphics[scale=1.0]{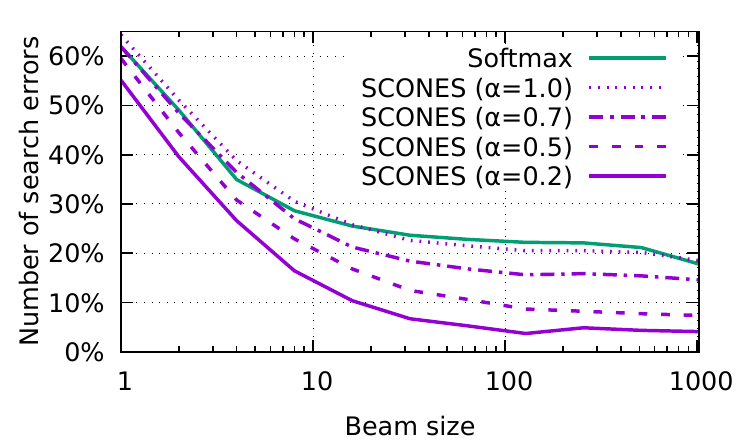}
\caption{Number of beam search errors for German-English as a function of the beam size.}
\label{fig:deen_beam_search_errors}
\end{figure}

Fig.\ \ref{fig:deen_beam_search_errors} displays the percentage of beam search errors, the fraction of sentences for which beam search did not find the global best translation, as a function of beam size. We confirm the findings of \citet{stahlberg-byrne-2019-nmt} for softmax models: the percentage of search errors remains at a relatively high level of around 20\% even for very large beam sizes. Increasing the beam size is most effective in reducing the number of search errors for SCONES with a small value of $\alpha$. However, a small $\alpha$ does not always yield the best overall BLEU score (Fig.\ \ref{fig:deen_beam_size_bleu}). Taken together, these observations provide an insight into model errors in NMT: If we describe the ``model error'' as the mismatch between the global most likely translation and an adequate translation (following \citet{stahlberg-byrne-2019-nmt}), a small $\alpha$ would simultaneously lead to both fewer search errors (Fig.\ \ref{fig:deen_beam_search_errors}) and fewer model errors (Tab.\ \ref{tab:beam-vs-dfs}). Counter-intuitively, however, BLEU scores peak at slightly higher $\alpha$-values (Tab. \ref{tab:alpha}). A more sophisticated notion of model errors and search errors is needed to understand the complex inherent biases of beam search for neural sequence-to-sequence models.

\section{Experiments with synthetic language pairs}

Our main motivation for SCONES is to equip the model to naturally represent intrinsic uncertainty, i.e.\ the existence of multiple correct target sentences for the same source sentence. To examine the characteristics of SCONES as a function of uncertainty, we generated synthetic language pairs that differ by the level of ambiguity. For this purpose, we trained an IBM-3 model \citep{brown-etal-1993-mathematics} on the German-English training data after subword segmentation using MGIZA \citep{gao-vogel-2008-parallel}. IBM-3 is a generative symbolic model that describes the translation process from one language into another with a generative story, and was popular for finding word alignments for statistical (phrase-based) machine translation \citep{smt}. The generative story consists of different steps such as distortion (word reordering), fertility (1:$n$ word mappings), and lexical translation (word-to-word translation) that describe the translation process. The parameters of IBM-3 define probability distributions for each step. In this work we do not use IBM-3 for finding word alignments. Instead, for the original German sentences we sample synthetic English-like translations from the model with different sampling temperatures to control the ambiguity levels of the translation task. A low sampling temperature generates sentence pairs that still capture some of the characteristics of MT such as word reorderings, but the mapping is mostly deterministic (i.e.\ the same source token is almost always translated to the same target token). A high temperature corresponds to more randomness, i.e.\ more intrinsic uncertainty. Appendix \ref{sec:ibm3} contains more details about sampling from IBM-3. We train NMT models using either softmax or SCONES on the synthetic corpora.

\begin{figure}[t!]
\centering
\small
\includegraphics[scale=1.0]{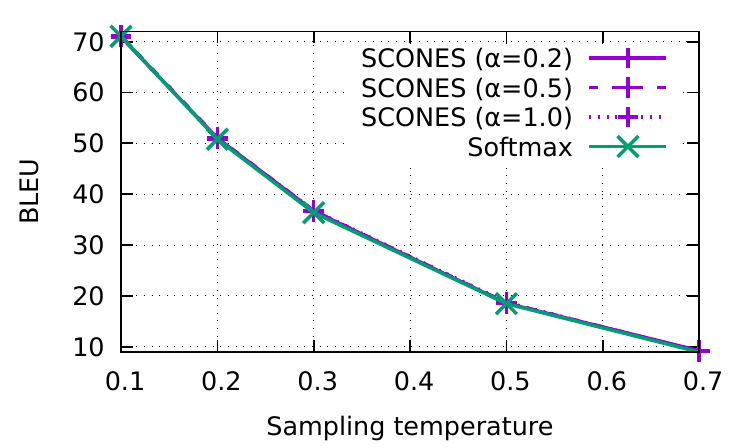}
\caption{BLEU scores with beam search (beam size of 4) for German-to-synthetic-English translation with different IBM-3 sampling temperatures.}
\label{fig:synth_beam4}
\end{figure}

\begin{figure}[t!]
\centering
\small
\includegraphics[scale=1.0]{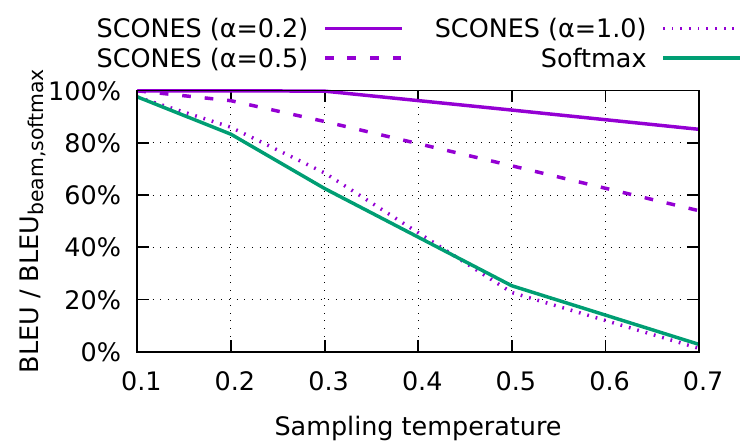}
\caption{Exact search for German-to-synthetic-English translation with different IBM-3 sampling temperatures. BLEU scores are shown relative to those of the beam search softmax output in Fig.\ \ref{fig:synth_beam4} for the respective temperatures.}
\label{fig:synth_dfs}
\end{figure}

Fig.\ \ref{fig:synth_beam4} shows that softmax and SCONES perform similarly using beam search: high IBM-3 sampling temperature translation tasks are less predictable, and thus lead to lower BLEU scores. The difference between both approaches becomes clear with exact search (Fig.\ \ref{fig:synth_dfs}). While the translations with the global highest probability for high IBM-3 sampling temperatures are heavily degraded for softmax and SCONES with $\alpha=1$, the drop is much less dramatic for SCONES with $\alpha=0.2$ (solid purple curve). Setting $\alpha$ to a low value enables the model to assign its highest probability to adequate translations, even when the translation task is highly uncertain.

\section{Related work}
Our approach draws insights from multi-label classification (MLC) \citep{tsoumakas07multilabelclassification,zhang-bp-mll,zhang14multilabel}. One of the earliest approaches for MLC was to transform the problem into multiple binary classification problems while ignoring the correlations between labels \citep{boutell2004multilabel}. More recent work has modeled MLC in the sequence-to-sequence framework with a decoder that generates the labels sequentially, thus preserving the inter-label correlations \citep{yang-etal-2018-sgm}. Most prior work in MLC focuses on classification and is not directly applicable to MT. In contrast, our training strategy is tailored for sequence-to-sequence problems. Unlike prior work \citep{yang-etal-2018-sgm}, SCONES allows us to perform MLC style training with any underlying NMT architecture by simply changing the loss function. By jointly
training all label-specific binary classifiers, our strategy is able to account for label correlations.

\citet{ma-etal-2018-bag} used an MLC objective to improve machine translation. Unlike our approach, they attempted to predict {\em all} words in the target sentence with a bag-of-words loss function. We formulate the next word prediction at each time step as an MLC problem to handle intrinsic uncertainty, but our models are predicting ordered target sequences, not bags of words.

The speed-ups from SCONES can be partially attributed to avoiding the normalization of the output over the full vocabulary. The same idea motivated earlier work on self-normalized training \citep{nce,nce-lm,devlin-etal-2014-fast,goldberger-melamud-2018-self}. As described in Sec.\ \ref{sec:scones}, unlike work on self-normalization, SCONES does not try to approximate a distribution over the full vocabulary. Rather, its output consists of multiple binary classifiers that do not share probability mass by design to be able to better represent intrinsic uncertainty.

\section{Conclusion}

Machine translation is a task with high intrinsic uncertainty: a source sentence can have multiple valid translations. We demonstrated that NMT models and specifically Transformers, can learn to model mutually non-exclusive target sentences from single-label training data using our SCONES loss function. Rather than learn a single distribution over all target sentences, SCONES learns multiple binary classifiers that indicate whether or not a target sentence is a valid translation of the source sentence. SCONES yields improved translation quality over conventional softmax-based models for six different translation directions, or (alternatively) speed-ups of up to 3.9x without any degradation in translation performance. We showed that SCONES can be tuned to mitigate the beam search curse and the problem of inadequate and empty modes in standard NMT. Our experiments on synthetic language translation suggest that, unlike softmax-trained models, SCONES models are able to assign their highest probability to adequate translations even when the underlying task is highly ambiguous.

The SCONES loss function is easy to implement. Adapting standard softmax-based sequence-to-sequence architectures such as Transformers requires \emph{only} replacing the cross-entropy loss function with SCONES and the softmax with sigmoid activations. The remaining parts of the training and inference pipelines can be kept unchanged. SCONES can be potentially useful in handling uncertainty for a variety of ambiguous NLP problems beyond translation, such as generation and dialog. We expect this work to encourage research on modeling techniques that can address ambiguity in much better ways compared to current models.


\bibliography{anthology,custom}

\clearpage

\appendix

\section{Time complexity of exact search}
\label{sec:dfs-maxedout}

The exact search algorithm of \citet{stahlberg-byrne-2019-nmt} we used in the paper is guaranteed to find the global best translation. Its runtime, however, varies greatly between language pairs and source sentences. Therefore, we limit the number of explored states per sentence by 1M to keep the decoding time under control. If the 1M threshold is reached, the optimality of the found translation is not guaranteed anymore. Fortunately, for most of our models and test sets, exact search was able to find and verify the global best translation earlier. Table \ref{tab:dfs-maxedout} lists the runs for which a fraction of the sentences did not terminate before 1M steps. In these rare cases, we use the best translation found thus far by exact search as an approximation to the global best translation.

\begin{table}[b!]
\centering
\small
\begin{tabular}{llc}
\hline \textbf{Languages} & \textbf{Run} & \textbf{\#incomplete sent.} \\
\hline
de-en & SCONES ($\alpha=0.2$) & 1.45\% \\
de-en & SCONES ($\alpha=0.5$) & 0.90\% \\
de-en & SCONES ($\alpha=0.7$) & 0.05\% \\
en-de & SCONES ($\alpha=0.2$) & 4.01\% \\
fi-en & SCONES ($\alpha=0.2$) & 1.30\% \\
en-fi & Softmax & 0.10\% \\
en-fi & SCONES ($\alpha=0.2$) & 3.20\% \\
lt-en & Softmax & 0.20\% \\
lt-en & SCONES ($\alpha=0.2$) & 5.20\% \\
en-lt & Softmax & 0.10\% \\
en-lt & SCONES ($\alpha=0.2$) & 5.31\% \\
synthetic-0.1 & Softmax & 1.05\% \\
synthetic-0.1 & SCONES ($\alpha=0.2$) & 0.55\% \\
synthetic-0.1 & SCONES ($\alpha=0.5$) & 1.10\% \\
synthetic-0.1 & SCONES ($\alpha=1.0$) & 1.25\% \\
synthetic-0.2 & Softmax	 & 1.00\% \\
synthetic-0.2 & SCONES ($\alpha=0.2$) & 5.10\% \\
synthetic-0.2 & SCONES ($\alpha=0.5$) & 7.65\% \\
synthetic-0.2 & SCONES ($\alpha=1.0$) & 2.65\% \\
synthetic-0.3 & Softmax	& 0.10\% \\
synthetic-0.3 & SCONES ($\alpha=0.2$) & 12.6\% \\
synthetic-0.3 & SCONES ($\alpha=0.5$) & 17.3\% \\
synthetic-0.3 & SCONES ($\alpha=1.0$) & 1.80\% \\
synthetic-0.5 & SCONES ($\alpha=0.2$) & 25.0\% \\
synthetic-0.5 & SCONES ($\alpha=0.5$) & 25.2\% \\
synthetic-0.7 & SCONES ($\alpha=0.2$) & 25.9\% \\
synthetic-0.7 & SCONES ($\alpha=0.5$) & 20.3\% \\
\hline
\end{tabular}
\caption{\label{tab:dfs-maxedout} Fraction of sentences for which exact search did not terminate before 1M steps. For runs that are not listed here, exact search terminated within 1M steps for all sentences.}
\end{table}

\section{Sampling from IBM-3}
\label{sec:ibm3}

The parameters of the IBM-3 model \citep{brown-etal-1993-mathematics} are composed of a set of fertility probabilities $n(\cdot|\cdot)$,
$p_0$, $p_1$, a set of translation probabilities $t(\cdot|\cdot)$, and a set of distortion probabilities $d(\cdot|\cdot)$. According to the IBM Model 3, the following generative process produces the target language sentence $\mathbf{y}$ from a source language sentence $\mathbf{x}$ \citep{knight-ibm}:

\begin{enumerate}
    \item For each source word $x_i$ indexed by $i = 1, 2, \dots, |\mathbf{x}|$, choose the fertility $\phi_i$ with probability $n(\phi_i
|x_i)$.
    \item Choose the number $\phi_0$ of ``spurious'' target words to be generated from $x_0 = \mathtt{NULL}$, using probability $p_1$ and
the sum of fertilities from step 1.
    \item Let $m=\sum_{i=0}^{|\mathbf{x}|} \phi_i$.
    \item For each $i = 0, 1, 2, \dots, |\mathbf{x}|$ and each $k = 1, 2, \dots, \phi_i$, choose a target word $\tau_{ik}$ with probability $t(\tau_{ik}|x_i)$.
    \item For each $i = 1, 2, \dots, |\mathbf{x}|$ and each $k = 1, 2, \dots, \phi_i$, choose a target position $\pi_{ik}$ with probability $d(\pi_{ik}|i, |\mathbf{x}|, m)$.
    \item For each $k = 1, 2, \dots, \phi_0$, choose  a position $\pi_{0k}$ from the $\phi_0-k+1$ remaining vacant positions in $1, 2, \dots, m$, for a total probability of $\frac{1}{\phi_0!}$.
    \item Output the target sentence with words $\tau_{ik}$ in positions $\pi_{ik}$ ($0 \leq i \leq |\mathbf{x}|$, $1 \leq k \leq \phi_i$).
\end{enumerate}

First, we estimate the IBM-3 model parameters using the MGIZA \citep{gao-vogel-2008-parallel} word alignment tool. Then, we sample English-like target sentences for the German source sentences following the generative story above. To control the level of uncertainty in the synthetic translation task we alter the entropies of the $n(\cdot|\cdot)$,  $t(\cdot|\cdot)$, and $d(\cdot|\cdot)$ distributions by choosing different sampling temperatures $\gamma\in\mathbb{R}^+$. Instead of sampling directly from a categorical distribution $P(\cdot)$ over categories $\mathcal{C}$, temperature sampling uses the following distribution:
\begin{equation}
    P_\gamma(c) = \frac{e^{\log P(c) / \gamma}}{\sum_{c'\in\mathcal{C}} e^{\log P(c') / \gamma}}
\end{equation}
for each $c\in\mathcal{C}$. A low temperature amplifies large differences in probabilities, and thus leads to a lower entropy and less ambiguity.

\section{Implementation of SCONES in JAX}
\label{sec:jax}

\begin{figure*}[t!]
\centering
\small
\begin{lstlisting}[language=Python]
from flax import linen as nn
import jax
import jax.numpy as jnp

def compute_scones_loss(
    logits,  # 3D float tensor [batch_size, max_sequence_length, vocab_size]
    targets,  # 2D int tensor [batch_size, max_sequence_length]
    l = 0.0,  # Label smoothing constant (lambda)
    a = 1.0,  # Scaling factor alpha
):
  true_logprob = nn.log_sigmoid(logits)
  false_logprob = jnp.log(jnp.maximum(1.0 - jnp.exp(true_logprob), 1.0e-30))
  gather = jax.vmap(jax.vmap(lambda s, t: s[t]))
  tgt_true_logprob = gather(true_logprob, targets)  # [batch_size, max_seq_length]
  tgt_false_logprob = gather(false_logprob, targets)  # [batch_size, max_seq_length]
  tgt_true_xent = -(1.0 - l) * tgt_true_logprob - l * tgt_false_logprob
  tgt_false_xent = -(1.0 - l) * tgt_false_logprob - l * tgt_true_logprob
  all_false_xent = -(1.0 - l) * false_logprob - l * true_logprob
  loss = a * (jnp.sum(all_false_xent, axis=-1) - tgt_false_xent) + tgt_true_xent
  weights = jnp.where(targets > 0, 1, 0).astype(jnp.float32)  # PAD ID is 0.
  return loss * weights / weights.sum()
\end{lstlisting}
\caption{JAX implementation of the SCONES loss function.}
\label{fig:jax}
\end{figure*}

Fig.\ \ref{fig:jax} provides an implementation of the SCONES loss function (Sec. \ref{sec:scones}) in JAX \citep{jax}. We bound the inverse model probability (\texttt{false\_logprob}) by $e^{-30}$ in line 12 for numerical stability. The JAX implementation generalizes the SCONES loss defined in the main paper in Eq.\ \ref{eq:scones} with a label smoothing \citep{ls} factor $\lambda\in[0,1]$ (\texttt{l} in Fig.\ \ref{fig:jax}) such that the positive loss component $\mathcal{L}_+(\cdot)$ becomes the following cross-entropy:
\begin{equation}
\begin{aligned}
\mathcal{L}_+(\mathbf{x},\mathbf{y}, i) =& -(1-\lambda)\log P(z_{\mathbf{x},\mathbf{y}_{\leq i}}=1|\mathbf{x}, \mathbf{y}_{< i}) \\
  & - \lambda \log P(z_{\mathbf{x},\mathbf{y}_{\leq i}}=0|\mathbf{x}, \mathbf{y}_{< i}).
\end{aligned}
\end{equation}
Similarly, the negative loss component $\mathcal{L}_-(\cdot)$ with label smoothing can be written as:
\begin{equation}
\begin{aligned}
\mathcal{L}_-(\mathbf{x},\mathbf{y}, i) =& -\sum_{w\in \mathcal{V}\setminus\{y_i\}} \big( \\
& (1-\lambda) \log P(z_{\mathbf{x},\mathbf{y}_{< i}\cdot w}=0|\mathbf{x}, \mathbf{y}_{< i}) \\
&+ \lambda\log P(z_{\mathbf{x},\mathbf{y}_{< i}\cdot w}=1|\mathbf{x}, \mathbf{y}_{< i}) \big). \\
\end{aligned}
\end{equation}
The label smoothing extension is provided for the sake of completeness -- we did not use label smoothing in any of the experiments in the main paper since it did not yield improvements in our setups.

\end{document}